\documentclass[11pt]{article}

\usepackage[a4paper,margin=25mm]{geometry}
\usepackage{fontspec}
\usepackage{xeCJK}
\usepackage{amsmath,amssymb}
\usepackage{graphicx}
\usepackage{booktabs}
\usepackage{array}
\usepackage{longtable}
\usepackage{tabularx}
\usepackage{pdflscape}
\usepackage{fancyvrb}
\usepackage{xcolor}
\usepackage{url}
\usepackage[hidelinks,unicode]{hyperref}

\setCJKsansfont{FandolHei-Regular.otf}[BoldFont=FandolHei-Bold.otf]
\setCJKmonofont{FandolFang-Regular.otf}

\hypersetup{
  pdftitle={From Expert Reduction to Behavioral Divergence: Tracing Numerical State through Sparse MoE Inference},
  pdfauthor={Tianyang Zhu}
}
\title{From Expert Reduction to Behavioral Divergence:\\
Tracing Numerical State through Sparse MoE Inference}
\author{Tianyang Zhu\\
\small Independent Researcher\\
\small \texttt{zty749@gmail.com}}
\date{}

\begin{document}
\maketitle

\begin{abstract}
Mathematically equivalent expert-reduction orders can produce observably different sparse-MoE executions. We isolate this effect in native DeepSeek-V4-Flash by freezing local MoE state and varying only aggregation semantics. Four schemes separate operand representation from accumulator precision. At one layer-5 fork, 720 A-mode orders yield 10 continuation basins; 720 B-mode orders form 360 exact structural classes and 11 basins. Under one Chinese prompt, the B classes split into 202 layoffs, 113 hiring, and 45 other continuations. Under maximum-$L_{\infty}$ B-branch selection, a 50-prompt exploratory breadth study finds cumulative text separation in 12, 24, and 36 prompts by 8, 16, and 32 tokens. Across the persistent trajectories, P32, A, and B change all 192 native-reference route trajectories per scheme; C preserves routes, token sequences, and texts. A separate 192-trajectory C check matches native MoE, post-mHC, next-router, and LM states bitwise. For one controlled B branch, exact post-mHC endpoint reconstruction reproduces the measured downstream trajectory. A complementary decode-boundary intervention reconstructs the branch’s complete attention-persistent state (hereafter, full persistent state) from production using an exact FP64 additive delta. With the naturally generated next input unchanged, 301 downstream post-mHC states, 301 full persistent state checkpoints, and 301 routes agree exactly over seven decode steps, as do predictions and text. These controls identify post-mHC as an intra-token state boundary and full persistent state as a cross-token continuation boundary. Identical emitted tokens need not imply identical autoregressive execution state: divergence can survive a token boundary and become visible later. Expert operand conversion, accumulator precision, and reduction order should form the numerical compatibility contract for sparse-MoE runtimes and hardware backends, not interchangeable kernel details. The results establish controlled causal possibility, not deployment incidence; C’s order invariance is limited to evaluated six-term states and schedules.
\end{abstract}


\section{Introduction}

Sparse MoE models reduce inference cost by routing each token to a small subset of experts. For a selected set of experts, the routed output is a weighted sum of expert contributions. The mathematical expression is permutation-invariant, but its floating-point implementation is not: different reduction orders can produce different finite-precision states even when the operands and gates are identical.

In a static feed-forward network, a small numerical difference may remain behaviorally irrelevant. A dynamic MoE decoder adds two discrete decision boundaries. The perturbed hidden state is consumed by later routers, whose top-k selection can change when a score margin is crossed. The resulting expert subgraph then feeds autoregressive greedy decoding, where a small logit difference can change the next token and all subsequent states. Thus, the reduction order of an expert aggregation is not necessarily a low-level implementation detail: it can become a hidden part of the model’s execution semantics.

Autoregressive execution also contains a persistent state boundary distinct from both routing and token selection. Even when two executions produce the same greedy token, numerical differences generated earlier in the forward pass can be written into layer-wise attention state and survive into later decode steps. The next token then starts from the same token embedding but consumes a different persistent execution state. This creates a mechanism for delayed output divergence: internal trajectories can remain different for several decode steps before their logit differences cross an argmax boundary.

Consequently, backends that implement identical model graphs and weights but differ in expert operand conversion, accumulator precision, or reduction topology need not be behaviorally compatible. Sparse-MoE backend qualification should therefore test not only final-token agreement but also persistent, layer-state, and routing conformance at internal state boundaries.

This paper addresses three research questions:

\begin{itemize}

\item \textbf{RQ1 — Reduction semantics and behavioral basins.} How do different same-mode reduction orders produce numerical, route, and continuation bifurcations? Event direction, horizon-dependent breadth, and deterministic replay provide complementary evidence for this question.

\item \textbf{RQ2 — Precision contract and compatibility.} How do operand representation and accumulator precision separately affect native compatibility and order sensitivity?

\item \textbf{RQ3 — State-boundary sufficiency and runtime conformance.} Which internal state boundaries are sufficient to reproduce a controlled divergent trajectory? In particular, is the post-mHC endpoint sufficient within a decode forward, and is the full persistent state endpoint sufficient to reproduce the branch across subsequent token boundaries when the next input token is unchanged?

\end{itemize}

We make three corresponding contributions:

\begin{enumerate}

\item We provide a trace-freeze-fork intervention that changes only expert aggregation semantics and maps the resulting post-mHC state space into continuation-text basins. A representative persistent trace separately observes a later routing boundary; event direction, breadth horizon, and exact replay characterize the behavioral structure of the branches.

\item We separate operand representation from accumulator precision using P32/C/A/B ablations and a same-mode canonical reference. C is stable over the evaluated states and schedules while preserving native-reference behavior.

\item We validate two state boundaries for one controlled branch. Exact post-mHC endpoint reconstruction reproduces the measured downstream intra-token trajectory, while exact reconstruction of the decode-boundary full persistent state reproduces the measured continuation across subsequent decode steps under the same naturally generated next input token. These controls motivate a hierarchical runtime-conformance procedure spanning persistent state, layer state, routing, and operator-level execution.

\end{enumerate}

The scope is deliberately limited. The experiments establish that the perturbation can produce routing and semantic bifurcations. They do not measure how often a real GPU, NPU, or distributed runtime naturally realizes each permutation. They also do not provide a frozen-route mediation control, an exact quantized-operand summation reference, or an end-to-end performance measurement.

\section{Background, related work, and methodology}

\subsection{Runtime, model, and decoding protocol}

All experiments use the native DeepSeek inference path in the Colibri project \cite{ref15}, without a Transformers reimplementation. The model name used in this paper is \textbf{DeepSeek-V4-Flash} \cite{ref12}. Prompts are passed as raw UTF-8 text, and decoding is greedy.

We use DeepSeek-V4-Flash with the native Colibri runtime because it provides implementation-level control and direct instrumentation of expert aggregation, layer states, routing, and persistent decode state. The investigation originated from instrumentation of mHC propagation in this runtime. A useful property of the checkpoint is that each token activates six routed experts, making all 6! = 720 cross-expert reduction orders tractable for exhaustive enumeration.

The project primarily supports the DSpark execution path \cite{ref14}. Because the standard checkpoint is evaluated through the non-DSpark path in this study, the command-line flag \nolinkurl{--no-dspark} is required as an explicit mode switch. It selects non-DSpark execution; it does not denote a different model name. This distinction is important because \nolinkurl{DeepSeek-V4-Flash} identifies the checkpoint/model, whereas DSpark and non-DSpark identify runtime execution paths.

All runs use one native-Windows CPU host. The runtime streams routed-expert weights from local storage and executes the evaluated path on the CPU. Full hardware, memory, compiler, build-flag, affinity, and explicit-thread replay details are reported in Appendix A.2.

The study uses two prompt groups. The depth cases are “why the sheep”, “朋友昨天打来电话”, and “Morning light filled the room”. The breadth group contains 25 English and 25 Chinese prompts and is used only as an exploratory check that the phenomenon is not confined to one prompt.

Persistent trajectories generate eight tokens. For each prompt and request seed, a deterministic layer-static permutation is selected independently for each MoE layer and then reused throughout that request. The four aggregation schemes share the same prompt, request seed, experiment seed, and layer-by-layer schedule. Thus 3 prompts \(\times\) 64 request seeds define 192 paired prompt-schedule conditions, each evaluated under four schemes:

\begin{equation*}
3\ \text{prompts}\times64\ \text{schedules per prompt}\times4\ \text{schemes}
=768\ \text{trajectories}.
\end{equation*}

Here a trajectory means one scheme-conditioned end-to-end eight-token greedy continuation with its route and text records, not one independent permutation or one token. All 768 candidate rows completed successfully and report eight generated tokens. The three production-reference baselines, one per prompt, are comparison runs and are not included in 768; counting them gives 771 native runs for the persistent protocol.

\subsection{Background and execution semantics}

\subsubsection{Sparse MoE execution}

The tested checkpoint routes each token to six routed experts and includes one shared expert. For selected expert i, let \(E_i(x)\) denote its expert output and \(g_i\) its router gate. The routed contribution is

\begin{equation*}
v_i = g_i E_i(x).
\end{equation*}

The routed contributions are reduced into a routed MoE output, after which the shared-expert output is merged. The resulting MoE output is passed to the rest of the block and is therefore part of the hidden-state trajectory, not a terminal statistic.

\subsubsection{mHC propagation, persistent state, and decision boundaries}

Within one decode forward, the mHC state carries the effect of an MoE perturbation into later layers. At attention boundaries, the resulting layer-wise states also update persistent execution state that is reused by later decode steps. This distinguishes three boundary types:

\begin{enumerate}

\item \textbf{Latent state boundary:} post-mHC state within the current forward.

\item \textbf{Persistent state boundary:} attention state retained across decode steps.

\item \textbf{Decision boundaries:} router top-k selection and LM-head argmax.

\end{enumerate}

A perturbation can cross the first two boundaries without crossing the token decision boundary. Thus identical emitted tokens do not imply identical persistent execution states. A route difference is also not automatically a token or semantic difference.

\subsubsection{Native reference aggregation path}

The controlled probe makes the rounding points explicit. The captured gate-weighted terms are represented as FP32 software values \(v_i\); their BF16 roundings are denoted by \(\widehat v_i=R_{\mathrm{BF16}}(v_i)\). Gate multiplication occurs before this controlled term-rounding step. The routed reduction is accumulated in an FP32 container, with scheme-specific rounding after each addition.

The shared-expert forward path returns a BF16-rounded output, which is added only after the routed sum. All four controlled schemes merge the same shared operand in the same fixed order, and the final merged output is converted with the runtime's BF16 rounding operation. Thus, the schemes vary only the routed-term and routed-accumulator semantics specified below.

The runtime implements \nolinkurl{R_BF16} in software. For finite FP32 values it adds the bit-level bias \nolinkurl{0x7fff + lsb_retained} and clears the low 16 mantissa bits, which is round-to-nearest, ties-to-even; this conversion does not depend on the host floating-point rounding mode. NaN and infinity skip the finite-value bias and have their low 16 bits cleared. FP32 accumulation uses componentwise C \nolinkurl{float} additions in the compiled native path. The cross-expert recurrence contains a standalone addition, so it has no cross-term fused multiply-add; gate multiplication has already occurred before the captured term. The tested build uses \nolinkurl{-O3 -march=x86-64-v3} and does not use \nolinkurl{-ffast-math}. FTZ/DAZ is not explicitly configured or recorded, and subnormal incidence was not instrumented. The structural B proof below is consequently restricted to the observed finite, non-NaN/non-infinite operands and the stated software BF16 rounding rule.

\subsubsection{Measurement hierarchy}

We report five layers of equivalence:

\begin{enumerate}

\item \textbf{Operator/local numerical state:} MoE aggregation output and related intermediates.

\item \textbf{Layer state:} post-attention and post-mHC states where instrumented.

\item \textbf{Persistent state:} full persistent state at a decode boundary.

\item \textbf{Discrete behavior:} router selections and greedy token IDs.

\item \textbf{Semantic behavior:} continuation text and event direction.

\end{enumerate}

Different internal routes can retain the same argmax token or later reconverge. Basin proportions are measured under the experimental permutation or seed measure; they are not normal model posterior probabilities and do not estimate real runtime incidence.

\subsection{Related work}

Recent LLM-inference studies show that numerically different execution environments can yield different outputs even under nominally deterministic decoding. Yuan et al. identify batch size, GPU count and version, and numerical precision as sources of inference nondeterminism \cite{ref1}. Chodavarapu and Xu show that KV-cached and cache-free autoregressive paths can diverge because they realize different low-precision accumulation paths \cite{ref2}. That result motivates attention to cross-token state. Our controlled experiment asks a complementary question: after a locally isolated expert-reduction fork, can the branch continuation be reconstructed solely by restoring the full persistent state endpoint while leaving subsequent execution unmodified? We use direct endpoint reconstruction rather than infer persistence only from cache-on/cache-off behavioral differences. At the software-system level, CRADLE compares multiple framework backends and localizes propagated inconsistencies \cite{ref8}, while NNSmith uses differential testing against reference backends to find deep-learning compiler bugs \cite{ref9}. These works motivate cross-environment auditing, but do not isolate cross-expert permutation while holding a local MoE state fixed.

For MoE models, MoQE studies the interaction between expert robustness and low-bit quantization \cite{ref3}, while value-and-structure alignment explicitly targets quantization-induced top-k routing changes \cite{ref4}. This literature establishes that small numerical perturbations can interact with sparse routing. Distributed MoE systems additionally make expert dispatch and combine first-class runtime operations: DeepSpeed-MoE develops optimized inference and expert-parallel execution \cite{ref10}, and DeepEP supplies low-precision all-to-all dispatch/combine kernels \cite{ref11}. Our intervention asks a narrower complementary question: with the selected experts, gates, expert outputs, shared output, and prefix state held fixed, how much behavior can be attributed to permutation-equivalent cross-expert reduction semantics alone?

A separate numerical-computing literature develops reproducible or exact summation through fixed accumulators, superaccumulators, and deterministic parallel reductions \cite{ref5,ref6,ref7}. We do not propose an exact summation algorithm. Instead, P32/C/A/B expose which operand and accumulator contract matches the native reference, and the B quotient removes a provable first-pair symmetry before branch enumeration. The contribution is therefore a controlled link from reduction semantics to MoE state, later routing where measured, and greedy continuation, rather than a new general-purpose reduction primitive.

The evaluated checkpoint and state boundary follow the DeepSeek-V4 technical report \cite{ref12} and the mHC architecture introduced by Xie et al. \cite{ref13}. DSpark is a speculative-decoding system \cite{ref14}, not a model variant; the experiments use the checkpoint through Colibri's native non-DSpark execution path \cite{ref15}. These sources define the model and software context, whereas the present work studies the numerical semantics of one runtime's cross-expert combine.

\subsection{Experimental protocols}

\subsubsection{Single-layer trace-freeze-fork protocol}

At the selected MoE position, the prompt prefix, token position, fork layer, selected expert IDs, gate values, six weighted expert terms, shared-expert output, and full persistent state are held fixed. Only the permutation or aggregation scheme at that one MoE aggregation changes. After the fork, downstream mHC propagation, attention, routing, and greedy decoding evolve normally. This protocol is used for the “why the sheep” study, the Chinese event-direction study, and the A/B permutation and equivalence-class studies.

For a permutation \(\pi\), the routed aggregation is evaluated as

\begin{align*}
a_0 &= +0,\\
a_{j+1} &= \operatorname{Accumulate}\!\left(a_j,v_{\pi_j}\right).
\end{align*}

\subsubsection{Post-mHC state-substitution control}

To test whether the downstream effect of the single-layer reduction intervention is fully captured by its post-mHC endpoint, we use one already catalogued “why the sheep” scheme-B class (class 135) at layer 5. At the fork, both the unmodified native-reference (production) post-mHC state \(H_P\) and the selected B post-mHC state \(H_B\) are captured after the normal BF16 round trip, and the endpoint perturbation is defined as

\begin{equation*}
\Delta H = H_B-H_P.
\end{equation*}

In a second process, reduction probing and forking are disabled. The runtime follows the native-reference path and performs a literal FP32 \texttt{H\_P += }\(\Delta H\) once, after FFN \nolinkurl{hc_post} and its BF16 round trip at the identical layer and decode position. Before addition, the runtime requires the native-reference state to match the captured \(H_P\) bitwise; after addition, it requires the reconstructed state to match \(H_B\) bitwise. It does not copy \(H_B\) over the result.

The decode-only fork occurs while consuming the first emitted token (token 477, position 3), not before that token is emitted from the prompt-prefill state. We generate nine driver tokens so that the first eight emitted tokens are each processed by a decode forward. From the injection boundary onward we compare every post-mHC state (339 vectors of 16,384 FP32 values). We also compare all 344 router selections across the eight decode forwards, generated token IDs, and generated text. The eight processed decode inputs yield 8 \(\times\) 43 = 344 route checkpoints. Because injection follows the layer-5 post-mHC endpoint of the first forward, the post-mHC comparison includes that endpoint and layers 6–42 plus seven complete forwards: 38 + 7 \(\times\) 43 = 339 checkpoints. Thus this is an endpoint state-substitution control for one single-layer branch, not a claim that an operand-level B reduction is equivalent to arbitrary hidden-state noise or that a persistent B schedule can be collapsed into one perturbation.

The endpoint-gated trace does not include layers preceding the layer-5 fork in its first decode forward. We therefore run a separate two-process pre-fork control at the same position: one process replays the B branch and the other uses probe-disabled native-reference execution. Trace mode records the post-mHC states at layers 0–4, before either execution reaches the intervention boundary, and compares their metadata, FP32 bytes, maximum absolute difference, and SHA-256 fingerprints.

\begin{figure}[t]

\centering

\includegraphics[width=\linewidth]{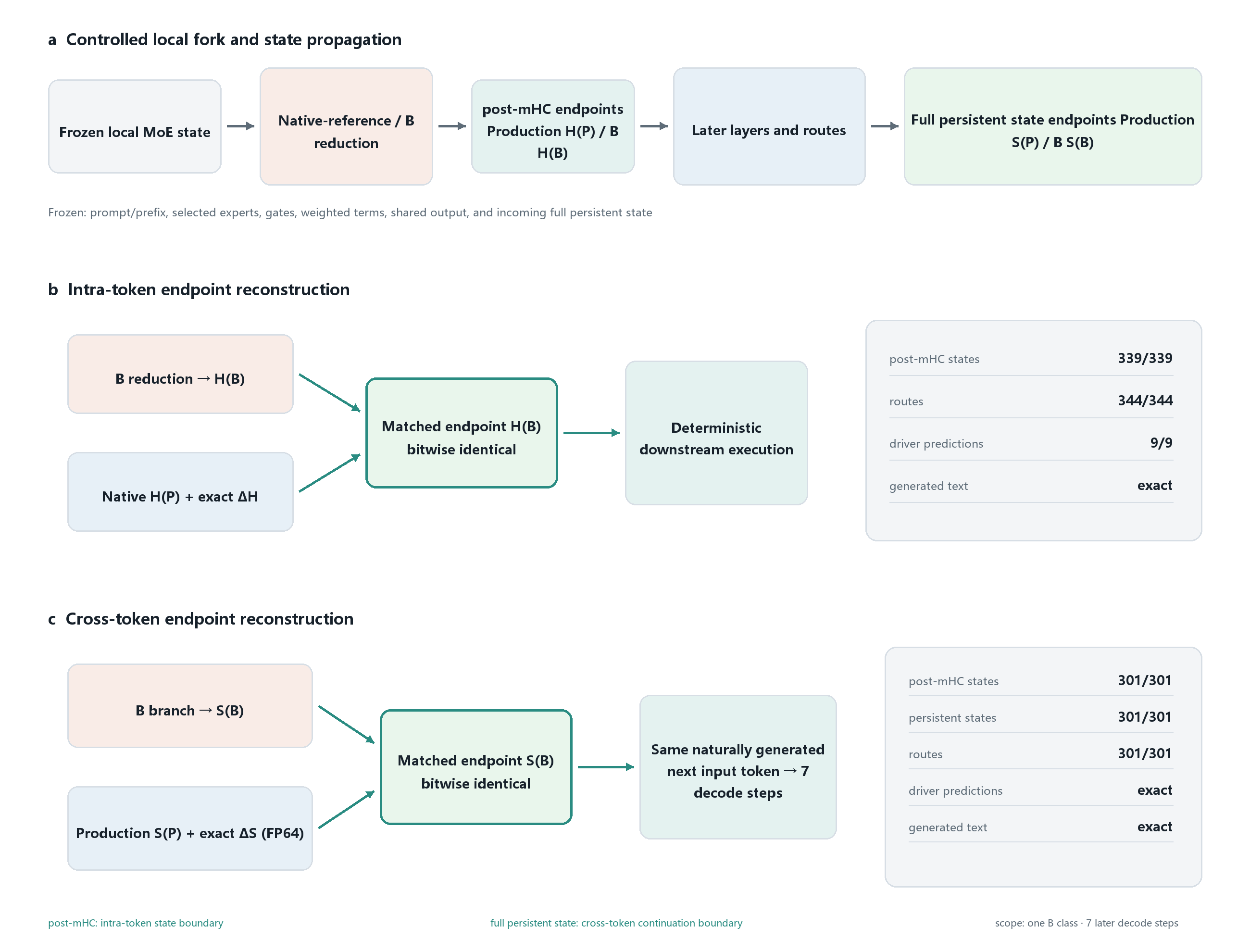}

\caption{Two complementary endpoint controls for one B class. Exact post-mHC reconstruction matches the branch endpoint and downstream trajectory, establishing an intra-token state boundary. At the following decode boundary, exact FP64 additive reconstruction of the full persistent state matches the branch endpoint; under the same independently generated next input token, 301/301 downstream post-mHC states, 301/301 full persistent state checkpoints, and 301/301 routes agree over seven later decode steps, as do the driver predictions and generated text. The control changes the endpoint, not subsequent execution.}

\end{figure}

\subsubsection{Decode-boundary persistent-state substitution control}

To test how the controlled reduction-path difference survives across token boundaries, we perform a second endpoint reconstruction on the same B class 135. The intervention is placed at the decode boundary after position 3 has completed all 43 layers.

The full persistent state is all floating-point runtime state retained by the attention subsystem for subsequent decode steps. In the evaluated runtime it comprises window KV, active compressed KV, compressor state, and indexer state, serialized in layer order under the same logical layout. Allocation padding and pointers are excluded. The position-3 endpoint contains 5,883,008 FP32 values. Let \(S_P\) and \(S_B\) denote the production and B-branch endpoints. We define

\begin{equation*}
\Delta S = S_B-S_P.
\end{equation*}

A third process follows production through position 3 and adds \(\Delta S\) to its full persistent state. Although both endpoints are FP32, subtracting them in FP32 can discard low bits needed to recover the target after subsequent addition: a literal FP32 delta reconstructs only 10/43 layer states bitwise. We therefore compute \(\Delta S\) in FP64 from the FP32 endpoints, add in FP64, and cast the reconstructed endpoint back to its runtime FP32 representation. This recovers all 43 layer states bitwise. The runtime requires the pre-add state to match \(S_P\) and the post-add state to match \(S_B\); it does not copy \(S_B\) into active state.

The FP64 delta is only an exact intervention representation. It is not a compact checkpoint format, and its success does not imply that the branch endpoint contains less information than a directly stored endpoint. Given \(S_P\), the exact delta is informationally sufficient to specify \(S_B\); the control tests endpoint sufficiency, not compression.

Injection occurs after position 3 has completed, so it cannot alter that forward or its already completed prediction. Branch and production must first independently select the same next driver token; no token is forced. Starting with position 4, seven subsequent decode inputs are executed while comparing every recorded post-mHC checkpoint, full persistent state checkpoint, router selection, predicted token ID, and generated text. The result concerns the full persistent state, not the most recent window-KV row alone. The seven post-injection decode inputs yield 7 \(\times\) 43 = 301 layer checkpoints for each post-mHC, full persistent state, and route comparison.

\subsubsection{Long-form event-direction extension}

For the Chinese prompt “朋友昨天打来电话”, we select ten scheme-B branches by deterministic stratified random sampling from the structural classes, using the recorded selection seed and drawing five branches from each provisional event direction. No qualitative content criterion is used to choose among eligible branches. The sampled fork is at layer 5, and each branch is extended from the original eight-token trace to 64 generated tokens. The rule-based coding used for sample selection assigned 202 classes to layoffs, 110 to hiring, and 48 to other continuations. Human consensus later corrected three \nolinkurl{other} classes to \nolinkurl{hiring}, yielding 202 layoffs, 113 hiring, and 45 other classes. The sample itself is a deterministic stratified sample, not an event-rate estimate.

The annotation units are the six unique exact eight-token continuations, not the 360 structural classes. Exact duplicate strings are collapsed before annotation, and the resulting labels are subsequently expanded to the classes by exact-text mapping through a recorded branch-to-item table. The blind form asks for \nolinkurl{hiring}, \nolinkurl{layoffs}, or \nolinkurl{other}, where \nolinkurl{other} includes indeterminate or incomplete continuations that do not explicitly establish either workforce direction. It omits branch IDs, branch frequencies, class multiplicities, and provisional labels. A disagreement would leave the consensus field blank pending adjudication; none occurred. The artifact retains two completed annotation columns but no annotator-identity or language-background metadata; we therefore do not claim that it independently verifies those attributes.

\subsubsection{Persistent layer-static protocol}

For each prompt and request seed, a deterministic permutation is selected independently for every MoE layer and reused at all token positions in that request. The four schemes share the prompt, request seed, experiment seed, and layer-by-layer schedule. Pairing controls the layer-wise permutation schedule, not the downstream expert operands after trajectories have diverged. After divergence, hidden states, selected experts, gates, and weighted terms are not shared. As defined in Section 2.1, the protocol has 192 paired prompt-schedule conditions and 768 scheme-conditioned eight-token trajectories. Three prompt-level native-reference baselines are additional and excluded.

\subsubsection{Same-mode canonical reference protocol}

The canonical reference fixes an identity permutation for each aggregation scheme and each of the three depth prompts. It separates a shift caused by changing the aggregation scheme from within-mode order sensitivity. The canonical experiment therefore contains 3 prompts \(\times\) 4 schemes = 12 eight-token trajectories.

{\small
\begin{longtable}{@{}p{0.130\linewidth}p{0.190\linewidth}p{0.210\linewidth}p{0.230\linewidth}p{0.180\linewidth}@{}}
\toprule
\textbf{Protocol} & \textbf{State scope} & \textbf{Intervention scope} & \textbf{Schedules or branches} & \textbf{Length} \\
\midrule
\endfirsthead
\toprule
\textbf{Protocol} & \textbf{State scope} & \textbf{Intervention scope} & \textbf{Schedules or branches} & \textbf{Length} \\
\midrule
\endhead
Single-layer permutation & One fixed token/layer state & One MoE aggregation & All A permutations / B structural classes & 8 tokens \\
Post-mHC substitution & One fixed B-class endpoint & One additive post-mHC perturbation & B class 135 versus native reference + exact endpoint delta & 8 processed decode tokens; 9 driver outputs \\
Persistent-state substitution & Complete position-3 decode boundary across 43 layers & Exact FP64 additive full-state reconstruction & B class 135 versus native reference + exact endpoint delta & 7 subsequent decode inputs; 9 driver outputs \\
Event-direction extension & One fixed Chinese fork & One MoE aggregation & 10 selected B classes & 64 tokens \\
Persistent intervention & Full request & Every MoE layer & 64 seeds/prompt; 192 paired prompt-schedules \(\times\) 4 schemes & 8 tokens \\
Canonical reference & Full request & Every MoE layer & Identity order & 8 tokens \\
\bottomrule
\end{longtable}
}

\subsubsection{Prompt and fork selection status}

The depth prompts and layer-5 fork are exploratory discovery cases; no prospective preregistration or held-out selection record exists for them. Layer 5 is hard-coded uniformly in the single-layer and breadth scripts and was not optimized separately for each prompt. \nolinkurl{FORK_TOKEN=-1} means the first eligible decode-only event is selected at runtime, not that a token ID was chosen in advance. The native logs report fork token IDs 477 for “why the sheep” and 303 for “朋友昨天打来电话”. Persistent and canonical experiments have no local fork because they apply their schedule at every MoE layer.

{\small
\begin{longtable}{@{}p{0.235\linewidth}p{0.235\linewidth}p{0.235\linewidth}p{0.235\linewidth}@{}}
\toprule
\textbf{Prompt set} & \textbf{Source and status} & \textbf{Single-layer fork} & \textbf{Persistent/canonical use} \\
\midrule
\endfirsthead
\toprule
\textbf{Prompt set} & \textbf{Source and status} & \textbf{Single-layer fork} & \textbf{Persistent/canonical use} \\
\midrule
\endhead
“why the sheep” & Exploratory depth case & Layer 5; logged fork token 477 & Included; no local fork \\
“朋友昨天打来电话” & Catalog ID 41; exploratory depth/event case & Layer 5; logged fork token 303 & Included; no local fork \\
“Morning light filled the room” & Catalog ID 4; exploratory depth case & Not used & Included; no local fork \\
50-prompt breadth set & Fixed \nolinkurl{prompt_breadth_50.tsv}; exploratory & Layer 5; first eligible decode-only event; inject maximum-$L_{\infty}$ B branch & Not used \\
\bottomrule
\end{longtable}
}

The breadth file was fixed as the batch input before the run and contains 25 English and 25 Chinese prompts: 40 ordinary natural prompts and 10 specially constructed sensitivity candidates. The script validates this composition, uses experiment seed 20260722, performs no adaptive prompt substitutions, and enumerates 360 B structural classes at each captured state before injecting the maximum-$L_{\infty}$ branch. The bilingual subsets were not matched by token count, so language counts are descriptive rather than a controlled cross-language comparison.

\subsection{Four aggregation schemes}

Let \(v_i\) = \(g_i\) \(E_i(x)\), let \(\widehat v_i=R_{\mathrm{BF16}}(v_i)\), and let \(R_{\mathrm{FP32}}\) denote the FP32 container update used by the controlled probe. For a permutation \(\pi\), the four schemes are:

{\small
\begin{longtable}{@{}p{0.235\linewidth}p{0.235\linewidth}p{0.235\linewidth}p{0.235\linewidth}@{}}
\toprule
\textbf{Scheme} & \textbf{Expert operand} & \textbf{Cross-expert accumulator} & \textbf{Purpose} \\
\midrule
\endfirsthead
\toprule
\textbf{Scheme} & \textbf{Expert operand} & \textbf{Cross-expert accumulator} & \textbf{Purpose} \\
\midrule
\endhead
P32 & FP32 terms & FP32 & Counterfactual higher-operand-precision control \\
C & BF16-rounded terms, exactly promoted to FP32 & FP32 & Native-like operands with protected accumulation \\
A & FP32 terms & BF16 round after every addition & Accumulator-precision control \\
B & BF16-rounded terms & BF16 round after every term and addition & BF16-native aggregation control \\
\bottomrule
\end{longtable}
}

More explicitly, with a zero initial accumulator:

\begin{align*}
\mathrm{P32}:\quad a_0 &= +0_{\mathrm{FP32}}, &
a_{j+1} &= R_{\mathrm{FP32}}\!\left(a_j+v_{\pi_j}\right),\\
\mathrm{C}:\quad a_0 &= +0_{\mathrm{FP32}}, &
a_{j+1} &= R_{\mathrm{FP32}}\!\left(a_j+
\mathrm{FP32}(\widehat v_{\pi_j})\right),\\
\mathrm{A}:\quad a_0 &= +0_{\mathrm{BF16}}, &
a_{j+1} &= R_{\mathrm{BF16}}\!\left(\mathrm{FP32}(a_j)+v_{\pi_j}\right),\\
\mathrm{B}:\quad a_0 &= +0_{\mathrm{BF16}}, &
a_{j+1} &= R_{\mathrm{BF16}}\!\left(\mathrm{FP32}(a_j)+
\mathrm{FP32}(\widehat v_{\pi_j})\right).
\end{align*}

The distinction between P32 and C is essential. C does not preserve the original FP32 expert terms; it first rounds each term to BF16 and then represents that BF16 value exactly in FP32 for accumulation. Scheme C is not introduced as a new exact-summation algorithm. It specifies a BF16-operand/FP32-accumulator execution contract that matches the native reference path over all evaluated states.

Let s denote the shared-expert output returned by the native forward path; s is already BF16-rounded and represented exactly in the FP32 software container. Every scheme applies the same shared-expert merge after the routed reduction:

\begin{equation*}
y_m=R_{\mathrm{BF16}}\!\left(a_6^m+\mathrm{FP32}(s)\right),
\qquad m\in\{\mathrm{P32},\mathrm{C},\mathrm{A},\mathrm{B}\}.
\end{equation*}

The B implementation redundantly applies \(R_{\mathrm{BF16}}\) to s before this merge. Because s is already BF16-rounded, that operation is bitwise idempotent and does not change the shared operand. The equations specify the controlled probe contract rather than a universal hardware floating-point semantics.

\subsection{Single-layer permutation and equivalence classes}

The single-layer intervention enumerates all 720 permutations for A. For B, the 720 raw permutations contain a structural symmetry. If the BF16-rounded terms are \(x_1,\ldots,x_6\) and \(x\oplus y\) denotes BF16-rounded addition, then

\begin{equation*}
0\oplus x_i=x_i.
\end{equation*}

and

\begin{equation*}
x_i\oplus x_j=x_j\oplus x_i.
\end{equation*}

Consequently, swapping the first two operands leaves the second-step accumulator bitwise unchanged, after which the suffix is identical. Under the tested finite operands, fixed BF16 rounding, componentwise addition, zero initialization, and no NaN/Inf or special signed-zero classification, B therefore has 720 / 2 = 360 exact structural equivalence classes. The experiment reports raw permutation count, structural class count, and observed unique MoE outputs separately. The class-weighted and raw-permutation-weighted basin proportions coincide because each class contains exactly two raw permutations and no additional collisions were observed.

\subsection{Native reference and same-mode canonical references}

We use two references for different questions. The native reference Y\_ref tests whether a replay preserves the behavior of the unmodified reference path. The same-mode canonical reference Y\_(m,\(\pi_0\)) fixes an identity permutation \(\pi_0\) for each aggregation scheme m. Within-mode order sensitivity is measured as

\begin{equation*}
D_{\mathrm{order}}(m,\pi)=D\!\left(Y_{m,\pi},Y_{m,\pi_0}\right).
\end{equation*}

This separates a shift caused by changing the aggregation scheme from a shift caused by changing the order within that scheme.

\subsection{Reproducibility and artifact scope}

The artifact includes scripts, raw-result indices, derived support tables, annotation records, and hash-based replay manifests; Appendix A provides the complete index.

The runtime records prompt IDs, modes, seeds, layer schedules, route traces, token IDs and hashes, generated text, exit codes, and log paths. The replay manifest additionally fixes the explicit thread configuration and records runtime, checkpoint, tokenizer, compiler, host, and affinity metadata. Raw logs, complete tensors, route traces, and model files remain external to the paper-facing support set.

Ordered token IDs are authoritative for continuation equality. Generated text is parsed as a marker-delimited multi-line record: an earlier single-line extractor truncated embedded newlines and undercounted extended-breadth text separation. The corrected Python analysis also joins staged runs by stable prompt ID so repeated later-horizon rows are not counted as new events. The breadth protocol disables direct next-router comparison and is consequently reported as post-mHC and text separation, not route divergence.

\section{Results}

\subsection{Native-reference replay correctness}

The native-reference replay check covers 13 prompts. Every replay preserves the reference route trajectory, greedy token sequence, and generated text, with c\_moe\_max\_abs = 0. This establishes that later differences are not caused by a broken prompt or replay harness. The unmodified native reference is therefore checked before interpreting any aggregation perturbation.

\subsection{A single reduction intervention creates state-to-text basins}

In the “why the sheep” single-layer experiment, A enumerates 720 raw permutations and produces 10 distinct continuation texts. B covers 720 raw permutations through 360 exact structural classes and produces 11 distinct texts. Every evaluated branch yields a post-mHC state different from the native reference: 720/720 A permutations and 360/360 B structural classes, with the same number of distinct post-mHC states. Direct next-router comparison was disabled in these sweeps, so this experiment does not assign a route-divergence label to every branch. The large numerical branch space nevertheless compresses into a much smaller set of text basins.

\begin{center}
\scriptsize
\resizebox{\linewidth}{!}{%
\begin{tabular}{@{}llllll@{}}
\toprule
\textbf{Scheme} & \textbf{Raw permutations} & \textbf{Exact structural classes} & \textbf{Distinct MoE outputs} & \textbf{Distinct post-mHC states} & \textbf{Distinct texts} \\
\midrule
A & 720 & 720 & 720 & 720 & 10 \\
B & 720 & 360 & 360 & 360 & 11 \\
\bottomrule
\end{tabular}%
}
\end{center}

Formally, for a branch set Π\_m and a text map f: Π\_m \(\rightarrow\) T, the basin of text t is B\_t = f⁻¹(t), and its reported volume is |B\_t| under the relevant permutation or class measure. In this sweep it is specifically post-mHC-state-to-text basin compression: many distinct internal states map to the same continuation. The persistent experiment below separately measures complete route trajectories. Basin volume is not a posterior probability, a geometric volume, or an estimate of deployment frequency.

\subsubsection{Representative persistent-schedule numerical-to-text propagation trace}

This subsection reports a separate persistent layer-static experiment, not the layer-5 single-layer fork used in Section 3.2.2. To display the propagation chain directly, we reran one already catalogued persistent condition: “why the sheep”, scheme B, request seed 0. The native reference and candidate use the same prompt and checkpoint. Scheme-B reduction semantics are active at every MoE layer under the deterministic layer-static schedule already represented in the persistent table; this is not a post hoc schedule search. At the last prompt position (position 2), the trace captures the layer-2 MoE and mHC outputs, the layer-3 router scores and selected experts, and the first LM logits. Route and token logging then continues through all eight generated tokens.

{\small
\begin{longtable}{@{}p{0.140\linewidth}p{0.250\linewidth}p{0.250\linewidth}p{0.300\linewidth}@{}}
\toprule
\textbf{Stage} & \textbf{Native reference} & \textbf{B, seed 0} & \textbf{Observed difference} \\
\midrule
\endfirsthead
\toprule
\textbf{Stage} & \textbf{Native reference} & \textbf{B, seed 0} & \textbf{Observed difference} \\
\midrule
\endhead
Layer-2 MoE output & Captured FP32 vector & Captured FP32 vector & Not bitwise equal; max \(L_{\infty}\) = 0.0375366211 \\
Layer-2 post-mHC state & Captured FP32 vector & Captured FP32 vector & Not bitwise equal; max \(L_{\infty}\) = 0.0307998657 \\
Layer-3 router & top-6 \{44, 93, 172, 174, 202, 225\}; rank-6/7 margin 0.00335884094 & top-6 \{44, 93, 143, 172, 174, 202\}; rank-6/7 margin 0.000484466553 & Score max \(L_{\infty}\) = 0.01445961; the selected-expert set changes by one entry: 143 enters and 225 leaves \\
First-position LM logits & argmax token 477; rank-1/2 margin 0.382600784 & argmax token 477; rank-1/2 margin 0.446601868 & Max \(L_{\infty}\) = 0.626680374, but the first greedy token is unchanged \\
First changed generated token & ordinal 5: token 11885 & ordinal 5: token 5477 & The first four generated token IDs agree; the fifth differs \\
Eight-token continuation & “are not in the fold. The sheep” & “are not in the pen. The sheep” & Different greedy continuation \\
\bottomrule
\end{longtable}
}

This representative trace exhibits the beginning and visible end of the propagation sequence: numerical MoE difference \(\rightarrow\) post-mHC difference \(\rightarrow\) later route difference \(\rightarrow\) delayed token/text difference. The persistent-state reconstruction in Section 3.2.3 supplies a complementary intervention on the previously unobserved cross-token link: latent divergence \(\rightarrow\) decode-boundary persistent-state divergence \(\rightarrow\) continued latent/route divergence. These boundaries must be reported separately: the layer-3 route changes before the first LM argmax, while the greedy sequence does not separate until generated-token ordinal 5. The trace does not establish that routing is the unique causal mediator, because no frozen-route control is performed.

\subsubsection{Exact post-mHC endpoint substitution reproduces the tested downstream trajectory}

For “why the sheep” B class 135, the native-reference state captured inside the probe run matches the state from the probe-disabled native-reference run bitwise at layer 5, position 3. Literal FP32 addition of the stored endpoint difference \(\Delta H\), whose maximum absolute component is 0.001953125, reconstructs the B post-mHC endpoint bitwise; the maximum reconstruction error is zero across all 16,384 values.

In the separate pre-fork control, the B replay and probe-disabled native-reference replay are bitwise identical at all five preceding post-mHC checkpoints (layers 0–4, using zero-based layer indices). Each checkpoint contains 16,384 FP32 values, all five maximum absolute differences are zero, and the paired SHA-256 fingerprints agree. This measures rather than assumes that the two executions enter the layer-5 fork from the same earlier-layer trajectory in that decode forward.

After that boundary, the B-reduction run and the native-reference-plus-\(\Delta H\) run remain bitwise identical for all 339 recorded post-mHC vectors spanning the rest of the first decode forward and seven complete subsequent decode forwards. All 344 compared router selections, all nine driver token IDs, and generated text also agree exactly. Nine outputs are used only to process eight emitted tokens; the ninth is the prediction produced after processing the eighth token.

{\small
\begin{longtable}{@{}p{0.235\linewidth}p{0.235\linewidth}p{0.235\linewidth}p{0.235\linewidth}@{}}
\toprule
\textbf{Control segment} & \textbf{Compared executions} & \textbf{Measured coverage} & \textbf{Result} \\
\midrule
\endfirsthead
\toprule
\textbf{Control segment} & \textbf{Compared executions} & \textbf{Measured coverage} & \textbf{Result} \\
\midrule
\endhead
Pre-fork trajectory & B replay versus probe-disabled native reference & Position 3, layers 0–4 post-mHC & 5/5 vectors bitwise identical; maximum \(L_{\infty}\) = 0 \\
Intervention endpoint & Captured B endpoint versus native \(H_P\) + \(\Delta H\) & Position 3, layer 5 post-mHC & 16,384/16,384 FP32 values bitwise identical; reconstruction error = 0 \\
Post-boundary latent trajectory & B branch versus native reference + \(\Delta H\) & Layer-5 endpoint, layers 6–42, and seven complete later forwards & 339/339 post-mHC vectors bitwise identical; every maximum \(L_{\infty}\) = 0 \\
Routing and visible outputs & Same pair over the complete measured horizon & 344 route checkpoints, nine recorded driver predictions, and generated text & Exact agreement \\
\bottomrule
\end{longtable}
}

This verifies the expected state-equivalence claim for one fixed single-layer branch: once the same post-mHC endpoint is reached with the same prefix and matched full persistent state, the measured deterministic downstream execution is identical over the evaluated horizon. It does not establish equivalence for an injection at the prompt-last state before token 477, for an analytically propagated pre-round MoE delta, or for persistent B interventions that introduce new perturbations later.

\subsubsection{Exact decode-boundary persistent-state reconstruction reproduces the branch continuation}

At the decode boundary after position 3, the B-class and production full persistent endpoints differ in 131,444 of 5,883,008 FP32 values. All nonzero differences occur in the 37 downstream layers 6–42; layers 0–5 remain bitwise identical, and the maximum absolute component difference is 1.0. This topology is consistent with the layer-5 fork occurring after that layer’s attention state has already been committed.

A literal FP32 encoding of \(\Delta S\) reconstructs only 10/43 layer records bitwise. In contrast, the FP64-encoded difference reconstructs all 43 layer records bitwise. The injected runtime independently verifies exact production and branch endpoints before and after addition; the captured B endpoint is not copied into active state.

The branch and production runs predict the same next driver token at this boundary, without forcing it. From that shared token onward, the B run and the production-plus-\(\Delta S\) run are bitwise identical for all 301 recorded post-mHC states and all 301 subsequent full persistent state checkpoints over seven decode inputs. All 301 compared router selections, all nine recorded driver predictions, and the generated text also agree. The nine predictions comprise one prompt-prefill prediction, the position-3 prediction completed before injection, and seven predictions produced by the post-injection forwards. Only the last seven fall within the intervention’s causal horizon; all nine are reported as a whole-run consistency check. The unmodified production continuation differs from the branch continuation, confirming that the reconstructed trajectory is not merely the native-reference trajectory.

For this branch and measured horizon, the reconstructed full persistent state endpoint is continuation-equivalent to the B endpoint at the evaluated decode boundary, conditional on the independently matching next token and deterministic execution environment. This establishes endpoint sufficiency, not uniqueness of computation history: different prior computations could in principle reach the same endpoint. It does not show that \(\Delta S\) can change a prediction completed before injection, that FP32 deltas are generally reversible, or that a window-KV-only view is sufficient.

\subsection{Opposing event-direction continuations}

For “朋友昨天打来电话”, the B full-branch output contains 360 structural classes. The two annotation columns collected from the blind six-item form agree on all six unique continuations (6/6; Cohen’s kappa = 1.0). Given the very small item count, kappa is reported descriptively rather than as strong statistical evidence. The six consensus labels are then expanded by exact-text mapping to the 360 classes, assigning 202 to layoffs, 113 to hiring, and 45 to other. Those mapped class rows are basin-volume accounting, not 360 independent annotation units. No disagreement required adjudication. Relative to the provisional rule, branches 95, 119, and 244 move from \nolinkurl{other} to \nolinkurl{hiring}. The class counts are not model probabilities or real-hardware frequencies.

To test whether the direction persists beyond a short prefix, we randomly sampled five layoffs classes and five hiring classes within their respective direction strata, using selection seed 20260723. Thus, the ten branches were not chosen after inspecting their long-form stories. The branch intervention remained B mode at layer 5 with the original prompt and fork settings; only the generation limit was increased from 8 to 64 tokens.

{\small
\begin{longtable}{@{}p{0.080\linewidth}p{0.100\linewidth}p{0.400\linewidth}p{0.360\linewidth}@{}}
\toprule
\textbf{Branch} & \textbf{Direction} & \textbf{Chinese continuation} & \textbf{English translation} \\
\midrule
\endfirsthead
\toprule
\textbf{Branch} & \textbf{Direction} & \textbf{Chinese continuation} & \textbf{English translation} \\
\midrule
\endhead
270 & layoffs & ，说他们公司要裁员，他可能被裁掉，心里很烦。我劝他，现在经济不景气，很多公司都在裁员，你被裁了，正好可以休息一段时间，再找新的工作。他听了我的话，心情好多了。 & ..., saying that his company was going to lay people off and that he might lose his job. He was very upset. I advised him that the economy was weak and many companies were laying people off; if it happened, he could take a break and then find another job. He felt much better after hearing this. \\
243 & hiring & ，说他们公司要招人，问我要不要去。我还在考虑，毕竟现在的工作也还行，但那边给的待遇确实不错。“哦？什么公司？”“一家做人工智能的初创公司，老板是海归，技术挺厉害的。” & ..., saying that his company was hiring and asking whether I wanted to join. I was still considering it because my current job was fine, although the compensation there was attractive. “What company?” “An AI startup; the founder returned from overseas and is technically very strong.” \\
\bottomrule
\end{longtable}
}

All ten randomly sampled branches produced 64-token continuations. Each of the five layoffs samples retained explicit workforce-contraction language, and each of the five hiring samples retained explicit workforce-expansion language. All ten continuations are textually distinct; beyond the shared hiring or layoffs direction, each follows a distinct narrative development, with branch-specific dialogue, situational details, and emotional framing. Appendix B.1 reports all ten complete streamed continuations; the two bilingual rows above illustrate the randomly sampled set rather than additional annotation units.

These branches exhibit an event-polarity bifurcation under an underdetermined prompt: the same entity and time frame receive opposite workforce-direction continuations. The six-item annotation and 10-trajectory extension support the opposing event-direction interpretation under the controlled B-class measure; they are not a large-sample semantic evaluation.

\begin{figure}[t]

\centering

\includegraphics[width=\linewidth]{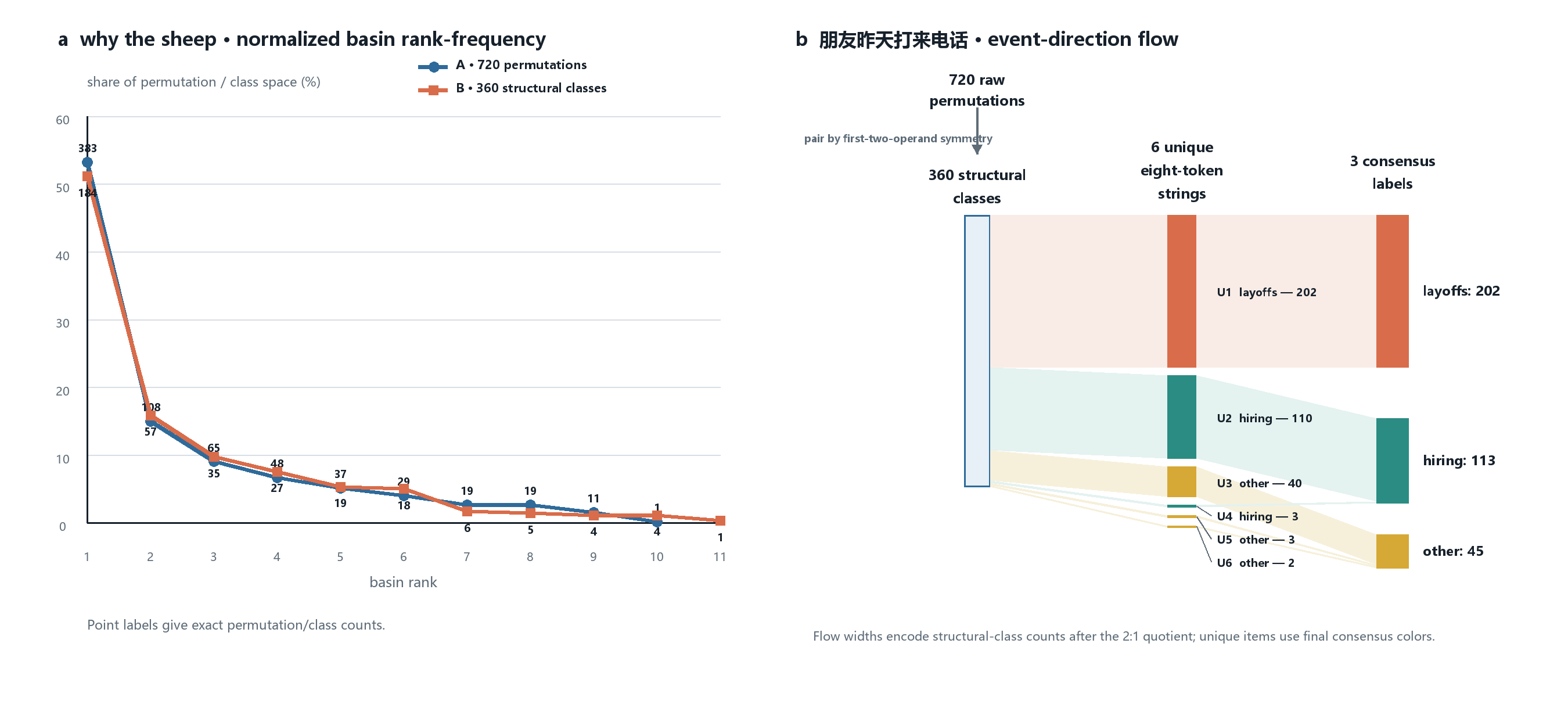}

\caption{Panel a compares normalized continuation-basin rank-frequency profiles for the “why the sheep” fork; point labels retain the exact A-permutation and B-structural-class counts. In panel b, the Chinese-prompt experiment first reduces 720 raw B permutations to 360 structural classes by the first-two-operand symmetry quotient. Alluvial widths begin only at the 360-class measure and continue through six unique eight-token strings to the final-consensus layoffs, hiring, and other volumes. Unique items are colored by their consensus labels, making the 110 + 3 = 113 hiring classes explicit.}

\end{figure}

\subsection{Deterministic replay of long-form branches}

To exclude run-to-run nondeterminism as an explanation for long-form narrative variation, we replayed the ten stratified-random scheme-B branches twice using identical prompt, fork layer, equivalence class, reduction schedule, model checkpoint, and runtime configuration. All 20 native runs completed successfully and generated 64 tokens. Each of the ten replay pairs produced an identical token-ID sequence and matching SHA-256 digest. Thus, the observed differences across long-form continuations are branch-conditioned and reproducible rather than consequences of uncontrolled decoding randomness. The comparison uses ordered token IDs and the SHA-256 hash of their canonical comma-joined sequence, which is stricter than text equality. Appendix B.2 gives the replay summary; Appendix A identifies the complete pair and environment metadata.

This establishes exact replayability for the tested 64-token horizon and the ten fixed branch conditions. It does not prove invariant behavior for the full 360-class space, arbitrary future checkpoints, changed hardware, changed thread counts or compilers, cross-model branch IDs, or all longer generation lengths.

\subsection{Exploratory breadth}

Across the fixed 50-prompt breadth set, every prompt has 360 distinct B post-mHC branch states and all 360 differ from the native-reference post-mHC state. Direct next-router comparison was disabled, so these 50/50 cases are post-mHC separation rather than measured route divergence. At eight tokens, at least two distinct continuation texts occur for 12/50 prompts. These cases include 6/25 English and 6/25 Chinese prompts, with 9/40 ordinary prompts and 3/10 specially constructed prompts.

After deduplicating repeated later-horizon rows by prompt ID, the cumulative result for the original fixed cohort is:

{\small
\begin{longtable}{@{}p{0.280\linewidth}p{0.670\linewidth}@{}}
\toprule
\textbf{Horizon} & \textbf{Cumulative separated} \\
\midrule
\endfirsthead
\toprule
\textbf{Horizon} & \textbf{Cumulative separated} \\
\midrule
\endhead
8 tokens & 12/50 \\
16 tokens & 24/50 \\
32 tokens & 36/50 \\
\bottomrule
\end{longtable}
}

Separation is horizon dependent in this staged exploratory cohort. The increasing cumulative count means that a longer greedy continuation reveals differences not visible at eight tokens; it is not a per-token probability estimate or a comparison of three independent cohorts. These counts remain an exploratory breadth check, not a systematic language comparison. Appendix B.3 reports stage membership, repeated rows, language/prompt-type breakdowns, and the multi-line parsing correction.

Future runtime studies could estimate horizon-dependent hazards with preregistered cohorts and survival analyses designed for survivor conditioning, unequal horizon intervals, and prompt heterogeneity.

\subsection{Persistent and same-mode reference axes}

The persistent experiment uses 64 paired layer-static schedules for each of three prompts and four schemes. All 768 scheme-conditioned trajectories exit successfully and generate eight tokens; the three native-reference baselines are separate. The same-mode canonical experiment adds one identity-order trajectory per prompt and scheme. Aggregating over prompts gives two reference axes:

{\small
\begin{longtable}{@{}p{0.188\linewidth}p{0.188\linewidth}p{0.188\linewidth}p{0.188\linewidth}p{0.188\linewidth}@{}}
\toprule
\textbf{Scheme} & \textbf{Native route} & \textbf{Native token sequence} & \textbf{Canonical route} & \textbf{Canonical token sequence} \\
\midrule
\endfirsthead
\toprule
\textbf{Scheme} & \textbf{Native route} & \textbf{Native token sequence} & \textbf{Canonical route} & \textbf{Canonical token sequence} \\
\midrule
\endhead
P32 & 192/192 & 161/192 & 192/192 & 115/192 \\
C & 0/192 & 0/192 & 0/192 & 0/192 \\
A & 192/192 & 178/192 & 192/192 & 123/192 \\
B & 192/192 & 172/192 & 192/192 & 114/192 \\
\bottomrule
\end{longtable}
}

The native columns compare random schedules with the production native reference; the canonical columns compare the same schedules with the identity-order trajectory of their own scheme. C is stable on both axes and is also native-compatible. P32, A, and B exhibit both a route/token-sequence shift from the native reference and within-mode order sensitivity. Route divergence is saturated for those three schemes, whereas token divergence is lower and not monotonically ordered, showing that changed route trajectories can reconverge to the same ordered token-ID sequence. Raw text is not used for this table because collection-only leading whitespace would create false differences. Appendix B.4 gives the per-prompt and canonical-baseline detail.

\begin{figure}[t]

\centering

\includegraphics[width=\linewidth]{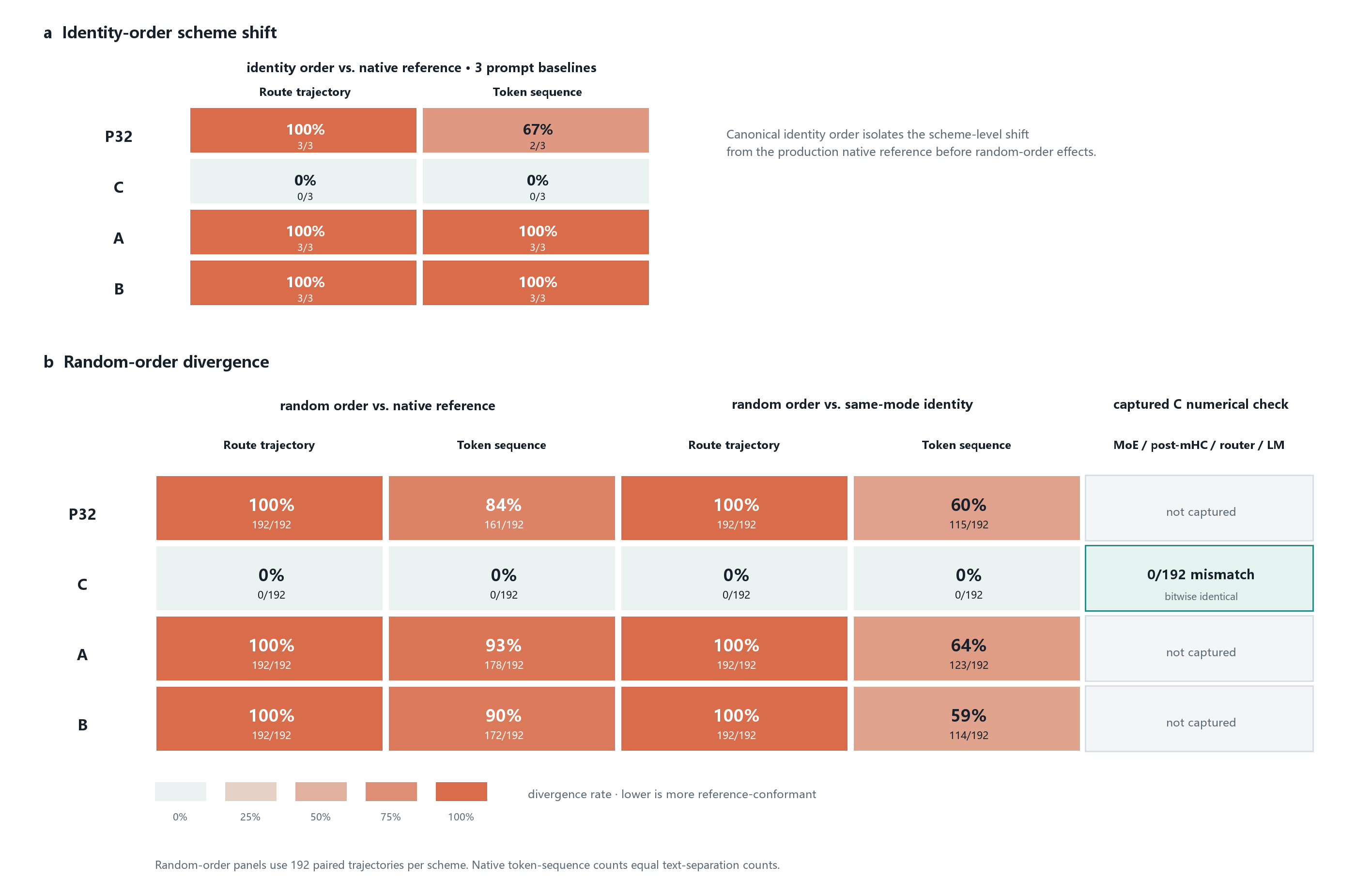}

\caption{Panel a reports the identity-order scheme shift against three production-native prompt baselines. Panel b reports random-order divergence against both the native reference and each scheme’s identity order over 192 paired trajectories per scheme; lower is better. The separately grouped C numerical check has 0/192 mismatches across captured MoE, post-mHC, next-router, and LM states. Intermediates were not captured for P32, A, or B.}

\end{figure}

The separate C intermediate check covers 3 prompts \(\times\) 64 schedules = 192 first-generation trajectories. The MoE aggregation output, post-mHC state, next-router scores, and prefill LM logits are bitwise identical to the native reference in every case, with zero missing records and maximum \(L_{\infty}\) distance 0. Together with the persistent and canonical results, this supports the scoped claim that C is empirically order-invariant while preserving the native-reference outputs for the evaluated six-term states and schedules. It does not establish exact summation, universal BF16 stability, or invariance under untested layers, states, schedules, or hardware topologies. Appendix B.5 gives the complete per-prompt intermediate summary.

\section{Discussion}

\subsection{Reduction order as execution semantics}

The results support a runtime-level interpretation of expert aggregation. The reduction order is mathematically invisible over real numbers but becomes observable when finite-precision aggregation changes a state near a router or token decision boundary. For a dynamic MoE decoder, a backend migration or kernel fusion can therefore alter the effective execution semantics even when the model graph and weights are unchanged.

\subsection{From local numerical divergence to delayed token divergence}

Together, the representative trace and two reconstruction controls establish the measured propagation sequence

\begin{center}
\small\sffamily
reduction semantics $\rightarrow$ intra-token latent divergence\\
$\rightarrow$ later layer/route divergence $\rightarrow$ decode-boundary persistent-state divergence\\
$\rightarrow$ cross-token latent/route divergence $\rightarrow$ delayed token difference
\end{center}

Routing is an observed discrete boundary but is not established as the unique causal mediator because no frozen-route intervention is performed. The persistent-state control addresses a different question: how a numerical divergence can survive a token boundary even when both executions emit the same next token. For the evaluated branch, reconstructing the full persistent state endpoint is sufficient to reproduce the measured continuation.

\subsection{Why P32 is not a ground truth}

P32 uses higher-precision expert operands, but that changes the values entering the aggregation. C preserves the native-reference BF16 operand semantics and changes only the cross-expert accumulator precision. The contrast between P32 and C shows that numerical compatibility is not obtained merely by increasing the precision of every intermediate. Semantic compatibility requires both operand semantics and accumulation semantics to be specified.

\subsection{Runtime and hardware implications}

The direct design implication is conditional. If several expert units produce partial results and a combine unit accumulates them in arrival order with low-precision sequential rounding, then completion order can become reduction order. This condition is distinct from a fixed software permutation and from a deterministic but hardware-specific reduction tree. A distributed or near-memory implementation can reduce this risk with BF16 expert compute plus an FP32 or guard-bit cross-expert accumulator, a fixed partial-sum tree, or deterministic merge semantics. Adaptive protection for small router margins is left as future work rather than treated as an evaluated intervention.

This experiment does not simulate analog compute noise, packet timing, or a specific CiM array. It controls the numerical semantics of the cross-expert merge and should therefore be presented as a design implication rather than a hardware measurement.

\subsection{From endpoint sufficiency to hierarchical conformance}

The two reconstruction controls identify complementary state boundaries. Post-mHC state provides a layer- and token-indexed boundary for local intra-forward diagnosis. Full persistent state provides a decode-boundary checkpoint for cross-token continuation. In the evaluated branch, exact reconstruction at either boundary reproduces the measured downstream trajectory from that boundary onward.

This suggests that runtime conformance need not begin with invasive operator-level tensor dumps. A candidate backend can first expose a canonical fingerprint of its full persistent state at the decode boundary before the divergent prediction. If that fingerprint differs from the reference, the difference can be bisected by persistent-state component and layer before enabling layer-indexed post-mHC tracing. The first post-mHC mismatch localizes the first observed state divergence to the interval after the last matched checkpoint; operator traces can then inspect the relevant attention, routing, expert, conversion, reduction, normalization, or residual computations in that interval. In the controlled experiments here, that search terminates at the expert-reduction operator.

{\small
\begin{longtable}{@{}p{0.080\linewidth}p{0.300\linewidth}p{0.560\linewidth}@{}}
\toprule
\textbf{Level} & \textbf{Compared object} & \textbf{Diagnostic purpose} \\
\midrule
\endfirsthead
\toprule
\textbf{Level} & \textbf{Compared object} & \textbf{Diagnostic purpose} \\
\midrule
\endhead
L0 & Token / text & Detect visible behavioral divergence \\
L1 & Logits / top-k margins & Detect pre-token decision divergence \\
L2 & Full persistent state & Detect cross-token state divergence \\
L3 & Layer-indexed post-mHC / post-attention state & Localize the first observed divergent layer \\
L4 & Relevant operator intermediates & Identify the divergent operator or numerical contract \\
\bottomrule
\end{longtable}
}

This hierarchy is not a claim that every token mismatch must originate in previously divergent persistent state. If the full persistent state and all other specified inputs entering the divergent token are identical, the first observed difference is generated inside the current forward and diagnosis can begin directly with layer-state comparison. Conversely, a persistent-state mismatch identifies a divergent decode boundary but not necessarily the earliest differing hidden intermediate.

Bitwise fingerprints require a canonical serialization contract specifying component schema, tensor identity, logical shape, dtype, element order, byte order, signed-zero handling, NaN canonicalization, token/layer coordinates, padding exclusion, and hash algorithm. The hashed record must also bind the model and checkpoint identity, tokenizer and prompt prefix, sequence position, and decoding configuration. For composite persistent state, the schema must additionally define the identity and order of window KV, compressed KV, compressor state, indexer state, and associated sequence metadata. A backend with a different internal layout must materialize the same canonical byte sequence before comparison.

A bitwise match does not prove that two backends used the same computation path; it establishes observational equivalence at a compared boundary. The present evidence validates endpoint sufficiency of post-mHC state and full persistent state for one single-layer B class in one native runtime and motivates, but does not validate, the proposed cross-backend procedure. Coverage across devices, tolerance rules for non-bitwise backends, and checkpoint density remain open validation questions.

\section{Limitations and future work}

Our experiments cover one checkpoint and one native runtime. DeepSeek-V4-Flash was chosen because the implementation provides the internal control required for exact intervention and replay; quantitative sensitivity should not be assumed to transfer unchanged to other MoE architectures or runtimes. The study covers six-term expert aggregation and a finite set of prompts, layers, states, and reduction schedules. It does not measure the incidence of these permutations on real GPU/NPU backends, perform frozen-route mediation, construct an exact quantized-operand reference, measure C’s runtime overhead, or test token-dynamic schedules.

The persistent-state reconstruction is demonstrated for one controlled branch and one decode boundary; it is not evidence that all numerical divergences are carried by the same persistent-state components. It establishes endpoint sufficiency, not uniqueness: distinct computation histories may converge to the same full persistent state endpoint. The FP64 additive delta is an exact intervention representation rather than a compact checkpoint encoding, so no compression or storage-efficiency claim is made. The full persistent state is a runtime-specific collection, not shorthand for conventional K/V tensors on other architectures, and the exact result is conditional on an independently matching next token.

The extended breadth result is a staged, selection-conditioned cohort rather than three independent 50-prompt experiments, so its cumulative counts are descriptive rather than probability estimates. The collection audit also exposes two practical failure modes for future runtime studies: generated text may span lines even when the token count is correct, and later horizon files may repeat earlier events. Robust tooling should preserve raw logs, use token IDs or marker-delimited multi-line extraction, key records by stable prompt IDs, and distinguish new from repeated divergences before computing cumulative counts.

\subsection{Claim ledger for deterministic replay}

{\small
\begin{longtable}{@{}p{0.180\linewidth}p{0.330\linewidth}p{0.430\linewidth}@{}}
\toprule
\textbf{Claim} & \textbf{Evidence} & \textbf{Scope limitation} \\
\midrule
\endfirsthead
\toprule
\textbf{Claim} & \textbf{Evidence} & \textbf{Scope limitation} \\
\midrule
\endhead
The tested branches can be replayed token-exactly. & 10 fixed branches \(\times\) 2 runs; 10/10 token-ID and SHA-256 pairs agree. & Current checkpoint, native runtime, hardware, and execution configuration. \\
Branch-to-branch narrative differences are not ordinary run-to-run randomness. & Each branch is stable under replay while the ten branch outputs differ. & Does not prove replay for every possible branch or bitwise replay across hardware. \\
\bottomrule
\end{longtable}
}

Accordingly, we do not claim that all 360 branches are necessarily reproducible, that changed hardware, thread counts, or compilers preserve bitwise output, that branch IDs remain valid across model versions, that all longer generations remain permanently deterministic, or that replay has established cross-platform determinism. The precise claim is exact replayability of ten representative branches, each run twice, in the same native runtime execution environment.

The event-direction artifact contains rule-based provisional labels, two completed annotation columns, and ten extended representatives. The released table verifies label completeness and agreement for six unique items, but does not retain annotator-identity or language-background metadata. The larger layoffs basin should be interpreted only as a larger basin under the controlled permutation measure. It is not a corpus frequency, posterior probability, or prediction of deployment behavior. The replay audit is likewise limited to exact repeatability of the tested 64-token trajectories.

\section{Conclusion}

Controlled changes to sparse-MoE expert reduction order can select different deterministic execution trajectories. In the tested DeepSeek-V4-Flash native runtime, single-layer post-mHC states compress into a smaller set of continuation-text basins, while the persistent experiment exhibits route and text trajectory divergence. The B-mode event-direction extension produces distinct hiring and layoffs continuations under one underdetermined prompt. The paired replay audit shows that these long-form differences are reproducible for ten stratified-random branches under the same native execution environment. Scheme C further matches the native-reference path at the evaluated numerical, route, and text levels, supporting BF16 operand semantics with FP32 accumulation as a scoped stabilization contract.

For the evaluated branch, the propagation chain can be followed from the controlled reduction through post-mHC state, later routing, and the full persistent state before becoming visible as a delayed token difference. Exact reconstruction at both the post-mHC and persistent-state boundaries reproduces the measured downstream trajectory. Post-mHC state is therefore a useful intra-token conformance boundary, whereas full persistent state provides a practical first-line boundary for cross-token runtime diagnosis.

These results motivate treating numerical conformance as a hierarchy of state-transition contracts rather than only an output-level test. Practical backend qualification can begin with decode-boundary persistent-state fingerprints and progressively localize disagreements through layer states, routing, and operator-level arithmetic. The proposed procedure remains subject to matched state and deterministic execution and is not cross-backend validated here.

For runtime and hardware designers, the practical implication is to specify cross-expert operand representation, accumulator precision, and merge ordering explicitly, and to validate backend migrations against numerical, routing, and token-level reference behavior.

The staged breadth extension further shows that cumulative text separation can increase substantially with generation horizon even after a short continuation remains identical. These findings are execution-environment specific and do not establish universal cross-platform determinism or real deployment frequencies.

\appendix

\section{Artifact and reproducibility manifest}

This appendix identifies the paper-facing artifacts needed to audit every reported count. Commands are run from the repository root with an explicit \nolinkurl{--model-path}; raw logs and model files remain outside the repository. The complete command lines and collection cautions are given in \nolinkurl{docs/experiments/v4-paper-reproducibility.md}.

\subsection{Protocol index}

{\small
\begin{longtable}{@{}p{0.130\linewidth}p{0.190\linewidth}p{0.210\linewidth}p{0.230\linewidth}p{0.180\linewidth}@{}}
\toprule
\textbf{Protocol} & \textbf{Prompt or selection input} & \textbf{Seed or schedule} & \textbf{Python entry point} & \textbf{Paper-facing output} \\
\midrule
\endfirsthead
\toprule
\textbf{Protocol} & \textbf{Prompt or selection input} & \textbf{Seed or schedule} & \textbf{Python entry point} & \textbf{Paper-facing output} \\
\midrule
\endhead
Native replay & 13 fixed replay prompts & Native order & Runtime replay audit & Section 3.1 counts \\
Single-layer A/B fork & \nolinkurl{why the sheep}, layer 5 & A: 720 permutations; B: 360 structural classes & \nolinkurl{tools/run_v4_branch_sweep.py} & \nolinkurl{basin_counts.tsv} \\
Post-mHC substitution & \nolinkurl{why the sheep}, B class 135, layer 5 & 5 pre-fork layer checkpoints; 8 processed decode inputs & \nolinkurl{tools/run_v4_mhc_equivalence.py} & \nolinkurl{mhc_equivalence_summary.tsv}, \nolinkurl{mhc_prefork_equivalence.tsv}, \nolinkurl{mhc_equivalence.tsv} \\
Persistent-state substitution & \nolinkurl{why the sheep}, B class 135, position 3 boundary & 7 subsequent decode inputs & \nolinkurl{tools/run_v4_kv_equivalence.py} & \nolinkurl{kv_equivalence_summary.tsv}, \nolinkurl{kv_equivalence_layers.tsv}, \nolinkurl{kv_equivalence.tsv} \\
Event direction & \nolinkurl{朋友昨天打来电话}, layer 5 & 10 selected B classes; two 64-token runs each & \nolinkurl{tools/run_v4_event_direction_replay.py} & \nolinkurl{event_direction_counts.tsv}, \nolinkurl{replay_branches.tsv}, \nolinkurl{replay_pairs.tsv} \\
Breadth & \nolinkurl{prompt_breadth_50.tsv} & Staged 8/16/32-token horizons & \nolinkurl{tools/run_v4_prompt_breadth.py}, \nolinkurl{tools/recompute_v4_breadth_text_stats.py} & \nolinkurl{breadth_summary.tsv} \\
Persistent & Three depth prompts & Seed 20260722; 64 paired layer-static schedules per prompt & \nolinkurl{tools/run_v4_persistent_random.py} & \nolinkurl{persistent_ablation.tsv} \\
Same-mode canonical & Three depth prompts & Identity order in each scheme & \nolinkurl{tools/run_v4_canonical_reference.py} & \nolinkurl{canonical_reference.tsv} \\
C intermediate & Three depth prompts & Seed 20260722; 64 C schedules per prompt & \nolinkurl{tools/run_v4_intermediate_check.py} & \nolinkurl{c_intermediate_summary.tsv}, \nolinkurl{propagation_trace.tsv} \\
\bottomrule
\end{longtable}
}

All paper-facing filenames in the last column are relative to \nolinkurl{docs/experiments/v4-paper-support/}. The support README defines their columns and maps each table to the corresponding result section. The principal schemas record protocol identifiers and denominators, divergence or bitwise-equality counts, \(L_{\infty}\) summaries where applicable, and immutable SHA-256 values for token sequences or mHC states. Full token sequences, tensors, route traces, long continuations, native logs, and model shards are intentionally excluded.

\subsection{Environment and identity}

{\small
\begin{longtable}{@{}p{0.280\linewidth}p{0.670\linewidth}@{}}
\toprule
\textbf{Component} & \textbf{Evaluated configuration} \\
\midrule
\endfirsthead
\toprule
\textbf{Component} & \textbf{Evaluated configuration} \\
\midrule
\endhead
CPU & AMD Ryzen AI MAX+ 395, 16 cores / 32 hardware threads \\
Installed GPU & AMD Radeon 8060S; not used by the evaluated runtime path \\
System memory & 128 GiB \\
Runtime memory option & \nolinkurl{--ram 32}; planner budget, not an OS-enforced RSS limit \\
Operating system & Windows 11 x64, NT 10.0.26200 \\
Toolchain & MSYS2 UCRT64 GCC 16.1, target \nolinkurl{x86_64-w64-mingw32} \\
Compile flags & \nolinkurl{-O3 -march=x86-64-v3 -fopenmp}; no fast-math flag \\
Evaluated execution & Native Windows CPU runtime; weights and compute streams remain on CPU \\
\bottomrule
\end{longtable}
}

General collection runs inherited the operating-system scheduler and did not pin threads. The deterministic 64-token replay alone fixed \nolinkurl{OMP_NUM_THREADS=32}, \nolinkurl{OMP_DYNAMIC=FALSE}, and \nolinkurl{PIPE_WORKERS=8}; these settings must not be inferred for the other protocols. \nolinkurl{replay_environment.tsv} records that replay configuration together with CPU/OS identity, affinity policy, runtime commit and binary SHA-256, checkpoint revision and manifest SHA-256, and tokenizer-file SHA-256 values. These hashes identify the tested environment without redistributing the checkpoint or tokenizer. Raw outputs may be placed in any external directory; their location is not part of the experimental semantics.

\subsection{Audit boundary}

The paper-support directory is sufficient to verify the published aggregate claims, but not to rerun native inference by itself. A full rerun additionally requires the checkpoint, compiled runtime, prompt and branch inputs, and an external output directory. The breadth audit treats the corrected multi-line reanalysis files as authoritative and deduplicates later-stage events by stable prompt ID. The Python claim-audit script checks the support tables and available root results before submission.

\subsection{Complete paper-facing file index}

Collection and verification entry points:

\begin{itemize}

\item \nolinkurl{tools/run_v4_branch_prompt.py}

\item \nolinkurl{tools/run_v4_branch_sweep.py}

\item \nolinkurl{tools/run_v4_prompt_breadth.py}

\item \nolinkurl{tools/recompute_v4_breadth_text_stats.py}

\item \nolinkurl{tools/run_v4_persistent_random.py}

\item \nolinkurl{tools/run_v4_canonical_reference.py}

\item \nolinkurl{tools/run_v4_intermediate_check.py}

\item \nolinkurl{tools/run_v4_mhc_equivalence.py}

\item \nolinkurl{tools/run_v4_kv_equivalence.py}

\item \nolinkurl{tools/run_v4_event_direction_confirmation.py}

\item \nolinkurl{tools/run_v4_event_direction_replay.py}

\item \nolinkurl{tools/apply_v4_event_direction_annotations.py}

\item \nolinkurl{tools/score_v4_event_direction_annotation.py}

\item \nolinkurl{tools/audit_v4_paper_claims.py}

\item \nolinkurl{tools/build_v4_paper_pdf.py}

\end{itemize}

Annotation, prompt, and raw-result indices used by the paper:

\begin{itemize}

\item \nolinkurl{prompt_breadth_50.tsv}

\item \nolinkurl{v4_event_direction_annotation.tsv}

\item \nolinkurl{v4_event_direction_annotation_unique_blind.csv}

\item \nolinkurl{v4_event_direction_annotation_unique_map.csv}

\item \nolinkurl{v4_event_direction_random5_64.tsv}

\item \nolinkurl{v4_prompt_breadth_50_B_recomputed.tsv}

\item \nolinkurl{v4_prompt_breadth_50_B_no_text_diff_16_recomputed.tsv}

\item \nolinkurl{v4_prompt_breadth_50_B_no_text_diff_32_recomputed.tsv}

\end{itemize}

Derived support tables, all under \nolinkurl{docs/experiments/v4-paper-support/}:

\begin{itemize}

\item \nolinkurl{basin_counts.tsv}

\item \nolinkurl{basin_rank_frequency.tsv}

\item \nolinkurl{breadth_summary.tsv}

\item \nolinkurl{c_intermediate_summary.tsv}

\item \nolinkurl{canonical_reference.tsv}

\item \nolinkurl{event_direction_counts.tsv}

\item \nolinkurl{event_direction_unique_continuations.tsv}

\item \nolinkurl{human_annotation_agreement.tsv}

\item \nolinkurl{human_annotation_confusion.tsv}

\item \nolinkurl{mhc_equivalence.tsv}

\item \nolinkurl{mhc_prefork_equivalence.tsv}

\item \nolinkurl{mhc_equivalence_summary.tsv}

\item \nolinkurl{kv_equivalence.tsv}

\item \nolinkurl{kv_equivalence_layers.tsv}

\item \nolinkurl{kv_equivalence_summary.tsv}

\item \nolinkurl{persistent_ablation.tsv}

\item \nolinkurl{propagation_trace.tsv}

\item \nolinkurl{replay_branches.tsv}

\item \nolinkurl{replay_environment.tsv}

\item \nolinkurl{replay_pairs.tsv}

\item \nolinkurl{README.md}

\end{itemize}

Protocol documentation:

\begin{itemize}

\item \nolinkurl{docs/experiments/v4-paper-reproducibility.md}

\item \nolinkurl{docs/experiments/v4-breadth-16-32-verification.md}

\end{itemize}

\section{Extended result tables}

\subsection{Complete 64-token event-direction continuations}

The main text shows one bilingual layoffs example and one bilingual hiring example. The complete set of ten stratified-random branch continuations follows. Each row is one completed 64-token generation; the direction label is the human consensus mapped from its exact eight-token prefix.

{\small
\begin{longtable}{@{}p{0.090\linewidth}p{0.120\linewidth}p{0.740\linewidth}@{}}
\toprule
\textbf{Branch} & \textbf{Direction} & \textbf{Complete streamed continuation} \\
\midrule
\endfirsthead
\toprule
\textbf{Branch} & \textbf{Direction} & \textbf{Complete streamed continuation} \\
\midrule
\endhead
230 & layoffs & ，说他们公司要裁员，他担心自己会被裁掉。我问他，你觉得自己会被裁掉吗？他说，我工作能力很强，应该不会被裁掉。我说，那你担心什么？他说，我担心万一被裁掉怎么办。我说，万一被裁掉，那就再找一份工作 \\
241 & layoffs & ，说他们公司要裁员，他担心自己会被裁掉，所以想让我帮他出出主意。我问他：“你平时工作表现怎么样？”他说：“还可以吧，就是有时候会迟到早退。”我说：“那你觉得公司会留一个经常迟到早退的员工吗？”他沉默了一会儿， \\
156 & layoffs & ，说他们公司要裁员，他担心自己会被裁掉。我问他，你担心什么？他说，我担心自己找不到工作。我说，你担心找不到工作，那你就去找工作啊。他说，我担心找不到工作，所以我不敢去找工作。我说，你不敢去找工作，那你就等着 \\
343 & layoffs & ，说他们公司要裁员，他担心自己会被裁掉。我问他，你觉得自己会被裁掉吗？他说，我觉得自己很危险。我问他，为什么？他说，我们公司是做传统媒体的，现在新媒体冲击很大，公司业绩不好，肯定要裁人。我问他，那你有没有想过 \\
270 & layoffs & ，说他们公司要裁员，他可能被裁掉，心里很烦。我劝他，现在经济不景气，很多公司都在裁员，你被裁了，正好可以休息一段时间，再找新的工作。他听了我的话，心情好多了。 \\
243 & hiring & ，说他们公司要招人，问我要不要去。我还在考虑，毕竟现在的工作也还行，但那边给的待遇确实不错。” “哦？什么公司？” “一家做人工智能的初创公司，老板是海归，技术挺厉害的。” \\
35 & hiring & ，说他们公司要招人，问我要不要过去。我还在考虑，毕竟现在的工作也还不错。” 林雪儿道：“那你自己考虑吧，反正你做什么决定，我都支持你。” 两人聊着天，不知不觉就到了林雪儿家楼下 \\
135 & hiring & ，说他们公司要招人，问我要不要过去。我还在考虑，毕竟现在的工作也还不错。” “哦，那你自己考虑吧。” 两人有一搭没一搭的聊着，不知不觉就到了下班时间。 林雪收拾 \\
277 & hiring & ，说他们公司要招人，问我要不要去。我考虑了一下，觉得还是去试试看。毕竟，现在的工作虽然稳定，但发展空间有限。” “嗯，有想法是好的。不过，你也要考虑清楚，不要轻易放弃现在的工作。” \\
322 & hiring & ，说他们公司要招人，问我要不要过去。我还在考虑，毕竟现在的工作也还不错。” “那挺好的，多一个选择总是好的。” 两人聊着，不知不觉就到了小区门口。 林知意停下脚步，看着 \\
\bottomrule
\end{longtable}
}

\subsection{Deterministic replay summary}

{\small
\begin{longtable}{@{}p{0.280\linewidth}p{0.670\linewidth}@{}}
\toprule
\textbf{Quantity} & \textbf{Result} \\
\midrule
\endfirsthead
\toprule
\textbf{Quantity} & \textbf{Result} \\
\midrule
\endhead
Selected branches & 10 \\
Runs per branch & 2 \\
Successful executions & 20/20 \\
Generated length & 64 tokens per run \\
Exact ordered token-ID agreement & 10/10 branch pairs \\
Token-sequence SHA-256 agreement & 10/10 branch pairs \\
Replay-metadata consistency & 10/10 branch pairs \\
\bottomrule
\end{longtable}
}

\begin{landscape}
\scriptsize
\begin{longtable}{@{}p{0.117\linewidth}p{0.117\linewidth}p{0.117\linewidth}p{0.117\linewidth}p{0.117\linewidth}p{0.117\linewidth}p{0.117\linewidth}p{0.117\linewidth}@{}}
\toprule
\textbf{Branch} & \textbf{Expected basin} & \textbf{Runs} & \textbf{Tokens/run} & \textbf{Token IDs exact} & \textbf{SHA-256 exact} & \textbf{Metadata consistent} & \textbf{Both exit zero} \\
\midrule
\endfirsthead
\toprule
\textbf{Branch} & \textbf{Expected basin} & \textbf{Runs} & \textbf{Tokens/run} & \textbf{Token IDs exact} & \textbf{SHA-256 exact} & \textbf{Metadata consistent} & \textbf{Both exit zero} \\
\midrule
\endhead
35 & hiring & 2 & 64 & yes & yes & yes & yes \\
135 & hiring & 2 & 64 & yes & yes & yes & yes \\
156 & layoffs & 2 & 64 & yes & yes & yes & yes \\
230 & layoffs & 2 & 64 & yes & yes & yes & yes \\
241 & layoffs & 2 & 64 & yes & yes & yes & yes \\
243 & hiring & 2 & 64 & yes & yes & yes & yes \\
270 & layoffs & 2 & 64 & yes & yes & yes & yes \\
277 & hiring & 2 & 64 & yes & yes & yes & yes \\
322 & hiring & 2 & 64 & yes & yes & yes & yes \\
343 & layoffs & 2 & 64 & yes & yes & yes & yes \\
\bottomrule
\end{longtable}
\end{landscape}

\nolinkurl{replay_pairs.tsv} gives the per-branch outcomes and immutable identifiers; \nolinkurl{replay_environment.tsv} gives the common explicit-thread environment. These results establish repeatability only for the ten randomly sampled branches under that environment, not for all 360 classes or across backends.

\subsection{Breadth stage accounting}

\begin{center}
\scriptsize
\resizebox{\linewidth}{!}{%
\begin{tabular}{@{}llllll@{}}
\toprule
\textbf{Horizon} & \textbf{At risk at stage start} & \textbf{New separated} & \textbf{Cumulative separated} & \textbf{Saved-file text differences} & \textbf{Repeated prior events} \\
\midrule
8 tokens & 50 & 12 & 12/50 & 12 & 0 \\
16 tokens & 38 & 12 & 24/50 & 12 & 0 \\
32 tokens & 26 & 12 & 36/50 & 17 & 5 \\
\bottomrule
\end{tabular}%
}
\end{center}

At eight tokens the 12 separated prompts comprise 6/25 English and 6/25 Chinese prompts, or 9/40 ordinary and 3/10 specially constructed prompts. The 16- and 32-token stages carry forward only survivors, so those stages are not independent 50-prompt cohorts. Five rows in the 32-token saved file were already separated at 16 tokens and are excluded from the new-event count. The initial single-line parser undercounted multi-line generations; the marker-delimited recomputation recovered 12 rather than 7 separations at 16 tokens and 17 rather than 3 saved-file differences at 32 tokens. All audited logs report the requested generated-token count.

\subsection{Persistent and canonical per-prompt detail}

{\small
\begin{longtable}{@{}p{0.188\linewidth}p{0.188\linewidth}p{0.188\linewidth}p{0.188\linewidth}p{0.188\linewidth}@{}}
\toprule
\textbf{Prompt} & \textbf{Scheme} & \textbf{Native route divergence} & \textbf{Native token-sequence divergence} & \textbf{Unique continuations} \\
\midrule
\endfirsthead
\toprule
\textbf{Prompt} & \textbf{Scheme} & \textbf{Native route divergence} & \textbf{Native token-sequence divergence} & \textbf{Unique continuations} \\
\midrule
\endhead
“why the sheep” & P32 & 64/64 & 59/64 & 4 \\
“why the sheep” & C & 0/64 & 0/64 & 1 \\
“why the sheep” & A & 64/64 & 58/64 & 8 \\
“why the sheep” & B & 64/64 & 61/64 & 10 \\
“朋友昨天打来电话” & P32 & 64/64 & 39/64 & 8 \\
“朋友昨天打来电话” & C & 0/64 & 0/64 & 1 \\
“朋友昨天打来电话” & A & 64/64 & 60/64 & 10 \\
“朋友昨天打来电话” & B & 64/64 & 54/64 & 8 \\
“Morning light filled the room” & P32 & 64/64 & 63/64 & 9 \\
“Morning light filled the room” & C & 0/64 & 0/64 & 1 \\
“Morning light filled the room” & A & 64/64 & 60/64 & 10 \\
“Morning light filled the room” & B & 64/64 & 57/64 & 9 \\
\bottomrule
\end{longtable}
}

{\small
\begin{longtable}{@{}p{0.188\linewidth}p{0.188\linewidth}p{0.188\linewidth}p{0.188\linewidth}p{0.188\linewidth}@{}}
\toprule
\textbf{Scheme} & \textbf{Canonical vs. native token sequence} & \textbf{Canonical vs. native route} & \textbf{Random vs. canonical token sequence} & \textbf{Random vs. canonical route} \\
\midrule
\endfirsthead
\toprule
\textbf{Scheme} & \textbf{Canonical vs. native token sequence} & \textbf{Canonical vs. native route} & \textbf{Random vs. canonical token sequence} & \textbf{Random vs. canonical route} \\
\midrule
\endhead
P32 & 2/3 & 3/3 & 115/192 & 192/192 \\
C & 0/3 & 0/3 & 0/192 & 0/192 \\
A & 3/3 & 3/3 & 123/192 & 192/192 \\
B & 3/3 & 3/3 & 114/192 & 192/192 \\
\bottomrule
\end{longtable}
}

\subsection{Complete C intermediate summary}

\begin{landscape}
\scriptsize
\begin{longtable}{@{}p{0.117\linewidth}p{0.117\linewidth}p{0.117\linewidth}p{0.117\linewidth}p{0.117\linewidth}p{0.117\linewidth}p{0.117\linewidth}p{0.117\linewidth}@{}}
\toprule
\textbf{Prompt} & \textbf{Schedules} & \textbf{Missing records} & \textbf{MoE max \(L_{\infty}\)} & \textbf{post-mHC max \(L_{\infty}\)} & \textbf{Router max \(L_{\infty}\)} & \textbf{LM max \(L_{\infty}\)} & \textbf{All captured values bitwise identical} \\
\midrule
\endfirsthead
\toprule
\textbf{Prompt} & \textbf{Schedules} & \textbf{Missing records} & \textbf{MoE max \(L_{\infty}\)} & \textbf{post-mHC max \(L_{\infty}\)} & \textbf{Router max \(L_{\infty}\)} & \textbf{LM max \(L_{\infty}\)} & \textbf{All captured values bitwise identical} \\
\midrule
\endhead
“why the sheep” & 64 & 0 & 0 & 0 & 0 & 0 & yes \\
“朋友昨天打来电话” & 64 & 0 & 0 & 0 & 0 & 0 & yes \\
“Morning light filled the room” & 64 & 0 & 0 & 0 & 0 & 0 & yes \\
\bottomrule
\end{longtable}
\end{landscape}

\end{document}